\documentclass[a4paper,twoside]{article}

\usepackage{epsfig} 
\usepackage{subcaption}
\usepackage{calc}
\usepackage{amssymb}
\usepackage{amstext}
\usepackage{amsmath}
\usepackage{amsthm}
\usepackage{multicol}
\usepackage{pslatex}
\usepackage{apalike}
\usepackage{algorithm2e}
\usepackage[bottom]{footmisc}
\usepackage[outdir=./]{epstopdf}
\usepackage{SCITEPRESS} 

\usepackage{graphicx}
\usepackage{pifont}

\begin{document}

\title{DiT-Head: High-Resolution Talking Head Synthesis using Diffusion Transformers}
\author{\authorname{Aaron Mir\sup{1}, Eduardo Alonso\sup{1} and Esther Mondragón\sup{1}}
\affiliation{\sup{1}Artificial Intelligence
Research Centre, Department of Computer Science, City, University of London, Northampton Square, EC1V 0HB, London, UK}
\email{\{E.Alonso\, E.Mondragon\}@city.ac.uk}}

\keywords{Talking Head Synthesis, Diffusion Transformers.}

\abstract{  We propose a novel talking head synthesis pipeline called "DiT-Head," which is based on diffusion transformers and uses audio as a condition to drive the denoising process of a diffusion model. Our method is scalable and can generalise to multiple identities while producing high-quality results. We train and evaluate our proposed approach and compare against existing methods of talking head synthesis. We show that our model can compete with these methods in terms of visual quality and lip-sync accuracy. Our results highlight the potential of our proposed approach to be used for a wide range of applications including virtual assistants, entertainment, and education. For a video demonstration of results and our user study, please refer to our supplementary material.}

\onecolumn \maketitle \normalsize \setcounter{footnote}{0} \vfill
\section{\uppercase{Introduction}}
\label{sec:introduction}

Talking head synthesis is a challenging task that aims to generate realistic and expressive faces that match the speech and identity of a given person. In recent years, there has been a growing interest in the development of talking head synthesis models due to their potential applications in media production, virtual avatars and online education. However, current state-of-the-art models struggle with generalising to unseen speakers and limited visual quality. Most existing methods focus on person-specific talking head synthesis \cite{chen2020talking,doukas2021headgan,guo2021ad,shen2022dfrf,zhang2022meta,ye2023geneface} and rely on expensive 3D structural representations or implicit neural rendering techniques to improve the performance under large pose changes, but these methods still have limitations in preserving the identity and expression fidelity. Moreover, most methods require a large amount of training data for each identity, which limits their generalisation ability. 

With the rapid speed of innovation in machine learning, we have better-performing models for image-based tasks such as Latent Diffusion Models (LDMs) \cite{rombach2021highresolution} that can generate novel images of high fidelity from text with minimal computational cost and Vision Transformers (ViTs) \cite{dosovitskiy2020image} that can capture global and local features of images and learn effectively from large-scale data. We argue that we should leverage these models to address the challenges of talking head synthesis in order to achieve higher fidelity and generalisation. We propose a novel talking head synthesis model based on Diffusion Transformers (DiTs) \cite{peebles2022scalable}
that takes a novel audio as a driving condition. Our model exploits the powerful cross-attention mechanism of transformers \cite{vaswani2017attention} to map audio to lip movement and is 
designed to improve the generalisation performance and the visual quality of synthesised videos. Additionally, DiTs are highly scalable, which can make them cost-efficient and time-efficient for a variety of tasks and dataset sizes.

The proposed model has the potential to significantly enhance the performance of synthesised talking heads in a wide range of 
applications, including virtual assistants, entertainment, and education. In this paper, we present the main methods and technical 
details of our model, as well as the experimental results and evaluation.

To summarise, we make the following contributions:
\begin{itemize}

  \item We design an LDM \cite{rombach2021highresolution} that substitutes the conventional UNet architecture \cite{ronneberger2015u} with a ViT \cite{dosovitskiy2020image,peebles2022scalable} that can scale up, handle multiple types of conditions and leverage the powerful cross attention mechanism of incorporating conditions. Audio is used to drive the denoising process and thus makes the talking head generation audio driven.
  \item Facial contour information is preserved through using a polygon-masked \cite{Ngo_2020_ACCV} ground-truth frame and a reference frame as additional conditions. This lets the network focus on learning audio-facial relations and not background information.
  \item The proposed DiT-Head model can generalise to multiple identities with high visual quality and quantitatively outperforms other methods for talking head synthesis in visual quality and lip-sync accuracy.
\end{itemize}

The remainder of the paper is organised as follows: \textbf{Section 2} reviews related work on talking head synthesis and explains the fundamentals of LDMs and DiTs; \textbf{Section 3} details our methodology; \textbf{Section 4} reports our results; \textbf{Section 5} discusses our findings, limitations, future work, and ethical issues, and \textbf{Section 6} concludes the paper.

\section{Related Work}

\paragraph{Talking Head Synthesis Models}

Talking head synthesis is the process of generating a realistic video of a person's face speaking, based on a driving input. This technology has many potential applications in areas such as virtual assistants, video conferencing, and entertainment. In recent years, machine learning models 
have been developed to improve the quality of audio-driven talking head synthesis. These models can be broadly divided into two categories: 2D-based and 3D-based methods.

2D-based methods \cite{zakharov2019few,zhou2019talking,chen2019hierarchical,zhou2020makelttalk,prajwal2020lip} use a sequence of 2D images of a person's face to synthesise a video of the person speaking a given audio. 
Wav2Lip \cite{prajwal2020lip} is a GAN-based talking head synthesis model that can generate lip-synced videos from audio. This method uses a lip-sync discriminator \cite{Chung16a} to ensure that the generated videos 
are accurately synchronised with the audio inputs. Wav2Lip uses an audio feature, a reference frame from the input video (a distant frame from the same video and identity), and a masked ground truth frame as inputs. 
This approach excels at producing accurate lip movements for any person, but the visual quality of the generated videos is suboptimal. 
This is because there is a trade-off between lip-sync accuracy and visual quality. Furthermore, GAN-based models suffer from mode collapse and unstable training \cite{wang2022talking}. More similar and concurrent to our work, DiffTalk \cite{shen2023difftalk} uses a UNet-based LDM to produce videos of a talking head. Smooth audio and landmark features are used to condition the denoising process of the model to produce temporally coherent talking head videos. 
% Wav2Lip consists 
% of two main components: a face detector during data pre-processing \cite{bulat2017far} and the lip-sync generator. 

3D-based methods \cite{chen2020talking,doukas2021headgan,guo2021ad,shen2022dfrf,zhang2022meta,ye2023geneface}, on the other hand, use a 3D model \cite{blanz1999morphable} of a person's face to synthesise a video of the person speaking the target audio. Meta Talk \cite{zhang2022meta} is a 3D-based method which uses a short target video to produce a high-definition, lip-synchronised talking head video driven by arbitrary audio. The target person’s face images are first broken down into 3D face model parameters including expression, geometry, illumination, etc. Then, an audio-to-expression transformation network is used to generate expression parameters. The expression of the target 3D model is then replaced 
and combined with additional face parameters to render a synthetic face. Finally, a neural rendering network \cite{ronneberger2015u,isola2017image}translates the synthetic face into a talking face without loss of definition. Other 3D-based methods use Neural Radiance Fields (NERFs) \cite{mildenhall2021nerf} for talking head synthesis \cite{guo2021ad,shen2022dfrf}.
NERFs for talking head synthesis work by modelling the 3D geometry and appearance of a person’s face and rendering it under different poses and expressions. They can produce more natural and realistic talking videos 
as they capture the fine details and lighting effects of the face and can handle large head rotations and novel viewpoints as they do not rely on 2D landmarks or warping \cite{guo2021ad,shen2022dfrf}.  

Both 2D-based and 3D-based methods have their advantages and disadvantages. 2D-based methods are computationally less expensive and can be trained on smaller datasets, but may not produce as realistic 
results as 3D-based methods \cite{wang2022talking}. 3D-based methods, on the other hand, can produce more realistic results, but are computationally more expensive and require larger datasets for training \cite{wang2022talking}. Furthermore, 3D-based methods rely heavily on identity-specific training, and thus do not generalise across different identities without further
fine-tuning \cite{guo2021ad,shen2022dfrf,wang2022talking}.

\paragraph{Latent Diffusion Models (LDMs)}

LDMs \cite{rombach2021highresolution} are a class of deep generative models that learn to generate high-dimensional 
data, such as images or videos \cite{yu2023video,blattmann2023align}, by iteratively diffusing noise through a series of transformation steps and 
then training a model to learn the reverse process. Given real high-dimensional data \(x\), an encoder \(E\) 
\cite{esser2021taming} can be used to learn a compressed representation of \(x\) to generate latent representation \(z\). Using maximum 
likelihood estimation, LDMs can learn complex latent distributions and generate high-quality samples from them \cite{rombach2021highresolution,lovelace2022latent,liu2023audioldm}. 

LDMs are versatile and can be applied to various tasks such as image and video synthesis, denoising, and inpainting. These type of generative model also have the advantage of having a single loss function and no discriminator, which makes them more stable and avoid GANs’ issues such as mode collapse and vanishing gradient \cite{wang2022talking}. Moreover, LDMs outperform GANs in sample quality and mode coverage \cite{rombach2021highresolution,dhariwal2021diffusion}. However, LDMs have a slow inference process because they need to iteratively run the reverse diffusion process on each sample to eliminate the noise.

\paragraph{Diffusion Transformers (DiTs)}

Diffusion Transformers (DiTs) are a recent type of generative model that blend the principles of LDMs and transformers. 
DiTs replace the commonly used UNet architecture in diffusion models with a ViT and can perform better than prior diffusion models on the class conditional ImageNet 512×512 and 256×256 benchmarks \cite{peebles2022scalable}. Given an input image, ViTs divide the image into patches and treat them as tokens for the transformer that are then linearly embedded. This allows the ViT to learn visual representations explicitly through cross-patch information interactions using the self-attention mechanism \cite{vaswani2017attention}.

Using transformers, DiTs model the posterior distribution over the latent space, which allows the model to capture intricate correlations between the latent variables. In contrast, standard LDMs use the UNet architecture \cite{ronneberger2015u} to model the same correlations. 
Transformers have some advantages over UNets in certain tasks that require modelling long-range contextual interactions and spatial dependencies 
\cite{cordonnier2019relationship}. They can leverage global interactions between encoder features and filter out non-semantic features 
in the decoder by using self- and cross-attention mechanisms \cite{vaswani2017attention,cordonnier2019relationship}. On the other hand, UNets are based on convolutional layers which are characterised by a limited receptive field and an equivariance property with respect to translations \cite{thome:hal-04058482}. In contrast, self-attention layers of ViTs allow handling of long-range dependencies with learned distance functions. Furthermore, ViTs are highly scalable which assists in handling a variety of image tasks of different complexities and diverse datasets. These powerful mechanisms make the ViT an intriguing architecture for use in multi-modal tasks such as talking head synthesis.

\paragraph{Conditioning Mechanisms of Diffusion Transformers}

Deep learning models often use conditioning mechanisms to incorporate additional information \cite{rombach2021highresolution,mirza2014conditional,cho2021unifying}. These mechanisms allow models to learn from multiple sources of information and can improve performance in many tasks. DiTs can incorporate 
conditional information in multiple ways and can model conditional distributions as \(q(z|c)\), where \(c\) represents the conditional information.

One such conditioning mechanism is concatenation, which involves appending the additional information to the input data. 
Although effective in natural language processing tasks like sentiment analysis, machine translation, and language modelling, concatenation is less useful when dealing with high-dimensional data of different modalities or unaligned conditional information. A more effective conditioning mechanism is that of cross-attention \cite{vaswani2017attention}. The cross-attention mechanism of transformers is a way of computing the relevance between two different sequences of embeddings. For example, in a vision-and-language task, the two sequences can be image patches and text tokens. Cross-attention calculates a 
weighted sum of the input data, where the weights depend on the similarity between the input data and the additional information. This mechanism has shown to be effective in multi-modal conditioning tasks e.g. text-to-image and text-to-video synthesis \cite{rombach2021highresolution,blattmann2023align}. 

The aim of this work was to produce a viable approach to high-quality, person-agnostic, audio-driven talking head video generation. We made use of the powerful self- and cross-attention mechanisms of DiTs in order to achieve this.

\section{Methodology}

\subsection{Data Pre-processing and Overview}

We employed a 2-stage training approach and an additional post-processing step at inference-time to produce temporally coherent lip movements using a DiT and incorporating audio features as a condition. For data pre-processing, we used a face-detection method \cite{zhang2017s3fd} to locate and crop the face from each frame of the input video. This face crop is then resized to \(H\times{W}\times{3}\) where \(H \in \mathbb{N}\) and \(W \in \mathbb{N}\). Next, facial landmark information was extracted from these images \cite{bulat2017far} where face images \(x \in \mathbb{R}^{H\times W\times3}\). Finally, a convex hull mask \cite{Ngo_2020_ACCV} was created over a copy of these images using facial landmark information for boundary definition where masked images \(x_m \in \mathbb{R}^{H\times W\times3} \). This design choice greatly impacted the quality of generated videos as it further localised the learning power of the self- and cross-attention mechanism used to learn the relation between the jaw, lips and audio and not the irrelevant information such as the neck, collar etc. Additionally, in order to assist in lip and lower-mouth generation and blending back into the ground-truth video, we performed Gaussian \(\alpha\)-blending to the polygon mask. i.e. the boundaries of the polygon mask were smoothed using a Gaussian kernel. For input audio pre-processing, we sampled at 16kHz and normalised in the range \(\left[-1,1\right]\).

In the first stage, two autoencoders were trained to faithfully reconstruct the ground-truth images and images where the mouth is masked. In the second stage, a DiT is trained whereby masked and reference frames are used in addition to an audio feature as conditions to drive the denoising process. The addition of masked and reference frames make the process more controllable and generalisable across different identities without requiring additional fine-tuning \cite{prajwal2020lip,shen2023difftalk}. Thanks to the learning mode of the latent space, the model can achieve high-resolution image synthesis with minimal computational costs.

In the next sections, we will provide details of the proposed DiT-Head pipeline including all stages. Figure 1 shows a visual overview of our architecture.

\subsection{Stage 1: Latent Feature Generation}

Taking our masked and ground-truth images, we trained two VQGANs \cite{esser2021taming} to retrieve the latent codes of the images with masks (\(E_{m}\)) and ground-truth images (\(E\)). Firstly, vector quantisation \cite{van2017neural} was applied to the continuous latent space which involves mapping continuous latent vectors to discrete indices, which are then used to represent the compressed latent space. The decoder for the unmasked images, (\(D\))
is only used for the final output and its weights and those of the encoders are fixed during DiT training. We reduced the dimensionality of the input face image by encoding it into a latent representation with a smaller \(H\) and \(W\) by a factor \(f\) where \(H/h = W/w = f\). This made the model learn faster and with less resources. The input face image \(x \in \mathbb{R}^{H\times W\times3}\) can be encoded into \(z_{gt} = E(x) \in \mathbb{R}^{h\times w\times3}\)  if it is unmasked, or \(z_{m} = E_{m}(x) \in \mathbb{R}^{h\times w\times3}\)
if it is masked.

\subsection{Stage 2: Conditional Diffusion Transformer}

In LDMs, a forward noising process is assumed which transforms sample \(z_{0}\) of the latent space to a noise vector through a series 
of \(T\) steps of diffusion. The diffusion process is defined by a Markov chain \cite{geyer1992practical} that starts from the data distribution 
\(q_0(z)\) and ends at a simple prior distribution \(q_T(z)\), such as a standard Gaussian. At each step \(t\), the latent variable \(z_{0}\) is 
corrupted by adding Gaussian noise \(\epsilon_t \sim \mathcal{N}(0, \beta_t I)\), where \(\beta_t\) is a noise level that increases with \(T\) steps \cite{ho2020denoising}. The full forward process can be formulated as,
\begin{equation}
\begin{aligned}
  q(z_{1:T} | z_{0}) := \prod\limits_{z=1}^T q(z_{t} | z_{t-1}),
\end{aligned}
\end{equation}

where the Gaussian noise transition distribution at each step \(t\) is given by:
\begin{equation}
\begin{aligned}
    q_t(z_t | z_{t-1}) := \mathcal{N}(z_t; \sqrt{1 - \beta_t} z_{t-1}, \beta_t I)
\end{aligned}
\end{equation}

where \(\beta_{t} \in (0,1)\) and \(1-\beta_{t}\) represent the hyperparameters of the noise scheduler. As defined by \cite{ho2020denoising}, using the rule of Bayes and the Markov assumption, we can write the latent variable \(z_{t}\) as:
\begin{equation}
\begin{aligned}
    q(z_t | z_{0}) := \mathcal{N}(z_t; \sqrt{\Bar{\alpha_{t}}}z_{0}, (1-\Bar{\alpha_{t}})I),
\end{aligned}
\end{equation}

where \(\Bar{\alpha_{t}} = \prod_{s=1}^{t}\alpha_{s}\), and \(\alpha_{t} = 1 - \beta_t\). Then, the reverse process \(q(z_{t-1} | z_{t})\) can be formulated as another Gaussian transition \cite{xu2023multimodal}:
\begin{equation}
\begin{aligned}
  q_{\theta}(z_{t-1}| z_{t}) := \mathcal{N} (z_{t-1};\mu_\theta(z_{t},t),\sigma_\theta(z_{t},t))),
\end{aligned}
\end{equation}

where \(\mu_\theta\) and \(\sigma_\theta\) denote parameters of a neural network \(\epsilon_{\theta}\). In our work, the DiT was used as \(\epsilon_{\theta}\) to learn the denoising objective such that we optimise:
\begin{equation}
\begin{aligned}
    L_{DiT} = \mathbb{E}_{z,\epsilon\sim N(0,1),t} \left[\| \epsilon - \epsilon_{\theta}\left(z_{t}, t\right) \|_2^2 \right]
\end{aligned}
\end{equation}

where the added noise \(\epsilon\) is predicted given a noised latent \(z_{t}\) where \(z_{t}\) is the result of applying forward diffusion to \(z_{0}\) and \(t \in [1,...,T]\). \(\tilde{z_{t}} = z_{t} - \epsilon_{\theta}(z_{t}, t)\) is the denoised version of \(z_{t}\) at step \(t\). The final denoised version \(\tilde{z_{0}}\) is then mapped to the pixel space with the pre-trained image decoder \(\tilde{x} = D(\tilde{z_{0}})\) where \(\tilde{x} \in \mathbb{R}^{H\times W\times3}\) is the restored face image.

Since we want to train a model that can create a realistic talking head video with the mouth in sync with the audio and the original identity is matching, audio features were used to condition the DiT.

\paragraph{Audio Encoding}

To incorporate audio information as a condition for talking head synthesis, we made use of a pre-trained Wav2Vec2 model \cite{baevski2020wav2vec}. Wav2Vec2 is a deep 
neural network model for speech recognition and processing, designed to learn representations directly from raw audio waveforms without requiring prior 
transcription or phonetic knowledge. The architecture is based on a combination of convolutional neural networks and transformers, and is 
trained using a self-supervised learning approach \cite{baevski2020wav2vec}.

In our approach, for a given input video, the pre-processed audio was converted to audio features using Wav2Vec2 \cite{baevski2020wav2vec}. From this audio and using the pre-trained model, individual features can be extracted for each frame of the video. We concatenated 11 audio features per frame to create an audio feature window of 11 (each frame at time-step \(t\) is associated with features from 5 frames prior and 5 frames after). Each concatenated feature window represents 0.44s of speech if the input video is sampled at 25 FPS. By concatenating audio information this way, we provided audio-visual temporal context. After introducing audio signals as a condition, the objective can be re-formulated as:
\begin{equation}
\begin{aligned}
  L_{DiT} = \mathbb{E}_{z,\epsilon\sim N(0,1),a,t} \left[\| \epsilon - \epsilon_{\theta}\left(z_{t}, a, t\right) \|_2^2 \right]
\end{aligned}
\end{equation}

where the concatenated audio features for the i-th frame are denoted as a 363-dimensional vector, \(a_{i} \in \mathbb{R}^{363}\). This audio feature is then further encoded using a linear transformation to \(a_{h} \in \mathbb{R}^{hidden}\) where \(hidden\) is the hidden dimension.

\paragraph{Person-agnostic Modelling}

Our goal was to create a model that can produce realistic lip movements for various people and speakers. To achieve this, we used a reference image as an input 
to our model, following the approach of Wav2Lip \cite{prajwal2020lip}. The reference image \(x_{r}\) contains information about the person’s face, head orientation and 
background. It is a random face image of the same person as the target frame, but from a different segment of the video (at least 60 frames away). We also used a 
masked image \(x_{m}\) as another input to guide the model for mouth region inpainting without relying on the real lip movements.

However, since DiT works in the latent space, we used \(E\) and \(E_{m}\) to retrieve the latent representations of \(x_{r}\) and 
\(x_{m}\) respectively where \(z_{r} = E\left(x_{r}\right) \in \mathbb{R}^{h\times w\times3}\) and \(z_{m} = E_{m}\left(x_{m}\right) \in \mathbb{R}^{h\times w\times3}\). Our final optimisation objective can be re-formulated as:
\begin{equation}
\begin{aligned}
  L_{DiT} = \mathbb{E}_{z,\epsilon\sim N(0,1), z_{t}, z_{r}, z_{m}, a, t} \left[\| \epsilon - \epsilon_{\theta}\left(z_{t}, z_{r}, z_{m}, a, t\right) \|_2^2 \right]
\end{aligned}
\end{equation}

\paragraph{Conditioning Implementation}

In order to integrate the conditioning information to the DiT, we used concatenation and cross-attention. We concatenated 
the spatially aligned \(z_{r}\)  and \(z_{m}\) and noised input latent \(z_{T}\) channel-wise. This was used as the input for the first layer of the DiT and introduced the masked ground-truth and reference frame as conditions to drive the inpainting process resulting in \(z_{T} \in \mathbb{R}^{h\times w\times9}\). Furthermore, the audio condition, \(a\) and time-step embedding, \(t\) were concatenated  channel-wise and then introduced to the DiT via cross-attention. In our model, the spatially aligned concatenated latents, \(z_{T}\) serve as the query, while the concatenated audio feature  \(a\) and time-step embedding \(t\) served as the key and value during cross-attention \cite{vaswani2017attention}. 
% Insert more information about cross-attention following LDM paper
\begin{figure*}[!ht]
  \centering
  \setlength{\belowcaptionskip}{-10pt}
  \includegraphics[width=0.95\textwidth]{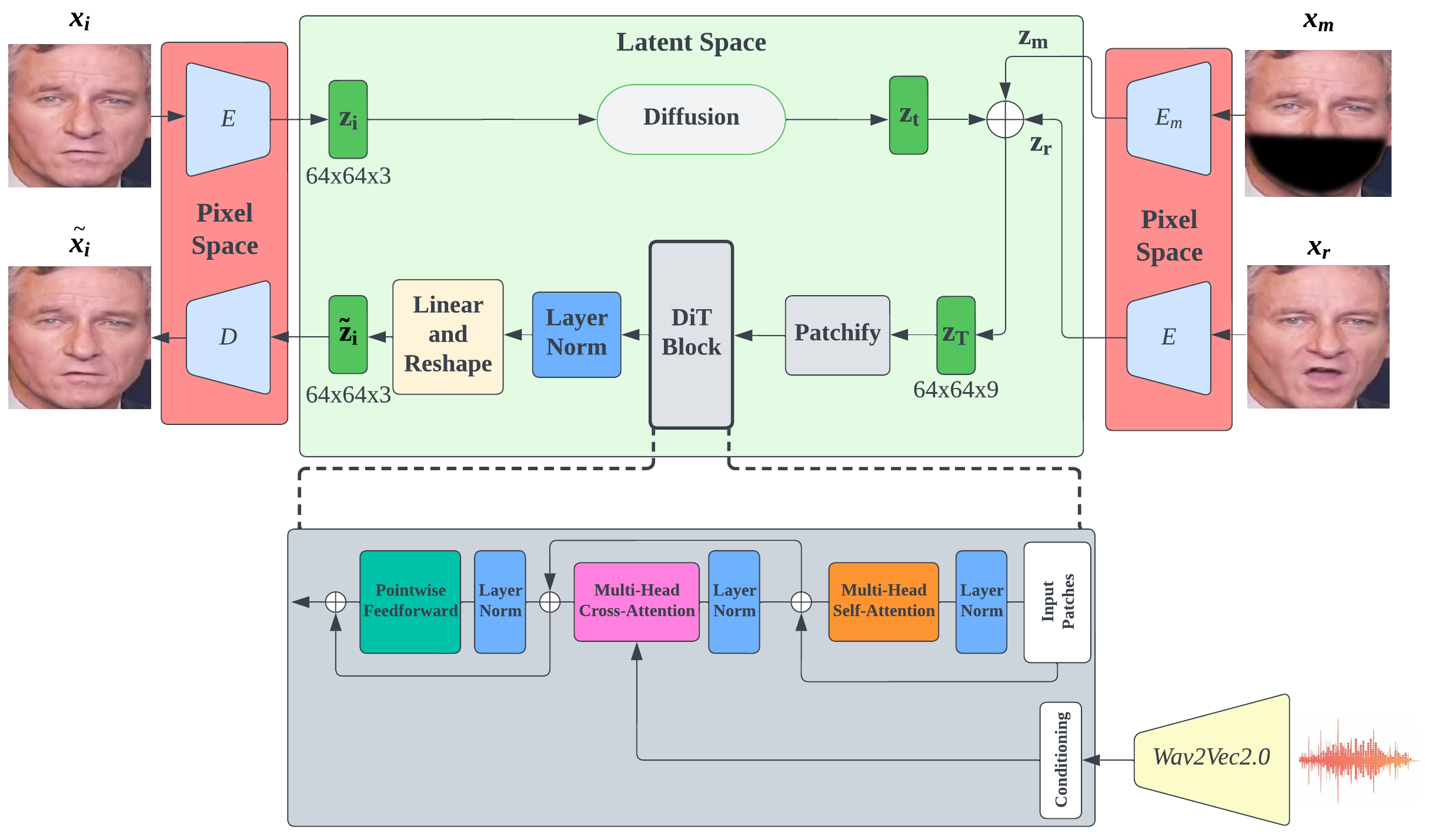}
\caption{\textbf{The DiT-Head architecture}. We train a diffusion-based model for generalised talking head video synthesis and use a DiT to learn the denoising process. The latent representations of the ground-truth image \(x_{i}\), reference image \(x_{r}\) and masked ground-truth \(x_{m}\) are extracted using VQGAN encoders \(E\) and \(E_{m}\). The latent representations of the reference image \(z_{r}\) and masked ground-truth \(z_{m}\) are concatenated with noise at time-step \(t\), \(z_{t}\), to produce intermediate representation \(z_{T}\). Additionally, we utilise the powerful cross-attention mechanism of transformers to introduce audio as a condition.}
\end{figure*}

\subsection{Evolving Inference and Output}

A Denoising Diffusion Implicit Model (DDIM) \cite{song2020denoising} was used as the sampling method to perform denoising during inference. This method of sampling is faster than the original Denoising Diffusion Probabilistic Model (DDPM) \cite{ho2020denoising} method. While both DDPM and DDIM are based on the diffusion process, they differ in their approach to modelling it. DDPM uses a neural network to model the data distribution and reduce noise step-by-step while DDIM learns the mapping from the data to the solution of a time-dependent partial differential equation that models the diffusion implicitly. Overall, DDIM can produce high-quality samples using less denoising steps \cite{song2020denoising}.

When generating a talking head video during inference, we provided the exact same conditioning as in training for the first frame (a masked image, a random reference image and a concatenated audio feature) however the ground-truth image latent \(z\) is substituted with noise. In addition, for subsequent frames, we use the generated noise latent at time-step \(t\) of the previous frame, \(z_{i-1}\), as the reference latent \(z_{r}\) as proposed by \cite{shen2023difftalk,bigioi2023speech}. In this way, the generated face of the previous frame is used to provide temporal contextual information for the generation of the next frame and results in a smoother evolution between frames at the output. During inference, we use 250 DDIM steps and the denoised output from the DiT for the i-th frame, \(\tilde{z_{i}} \in \mathbb{R}^{h\times w\times3}\), is decoded through the unmasked image decoder \(D\) to obtain output face image \(\tilde{x_{i}} = D(\tilde{z_{i}}) \in \mathbb{R}^{H\times W\times3}\).

\subsection{Stage 3: Post-processing}

We applied video frame-interpolation (VFI) on the resulting video in order to alleviate temporal jitter around the mouth region. This temporal jitter is common when 
attempting to iteratively generate temporally coherent images using diffusion-based models \cite{blattmann2023align}. VFI is a technique that can generate intermediate 
frames between two consecutive frames in a video, resulting in a smoother and more realistic motion. We performed 2x interpolation using the RIFE (Real-time Intermediate 
Flow Estimation) \cite{huang2020rife} model which uses a neural network that can directly estimate the intermediate flows from images, without relying on bi-directional optical flows that can cause artifacts on motion boundaries \cite{huang2020rife}.

\subsection{Training Details}

6 hours of randomly selected footage from the HDTF dataset \cite{zhang2021flow}, containing high-quality videos of diverse speakers, facial expressions and poses, was re-sampled to 25 FPS and used for training. We randomly shift the mask and landmarks by a few pixels during training to improve the model’s generalisation ability and use a Gaussian kernel of size \(27 \times{27}\) and \(\sigma\) of \(5\) for our mask. Our VQGANs \cite{esser2021taming} use a downsampling factor of \(f\) = \(4\), produce an intermediate embedding of \(z \in \mathbb{R}^{\frac{H}{f}\times{\frac{W}{f}}\times{3}}\) and use a learnable codebook of \(N_c \times{3}\). In our experiments, \(H\) = \(256\), \(W\) = \(256\), \(N_c\) = \(16384\) and our pre-trained Wav2Vec2 \cite{baevski2020wav2vec} model was \textit{wav2vec2-large-xlsr-53-english}. For the diffusion process, we use the same hyperparameters from ADM \cite{dhariwal2021diffusion}. Our learning rate was constant at \(1\times{10^{-4}}\) and our model had a patch size of 2, 24 DiT blocks, 16 attention heads and a hidden dimension of 1024 (DiT-L/2 \cite{peebles2022scalable}). The two VQGANs and DiT were trained for 72 hours each using 4 A100 GPUs.

\section{Results}

\subsection{Evaluation Details}

Our model was evaluated quantitatively on 11 unseen identities (around 18 minutes of video at 25 FPS) from our randomly selected HDTF \cite{zhang2021flow} test set using ground-truth driving audio. Peak Signal-to-Noise Ratio (PSNR) \cite{kotevski2010experimental}, Structural Similarity Index Measure (SSIM) \cite{wang2004image}, Learned Perceptual Image Patch Similarity (LPIPS) \cite{zhang2018perceptual}, Fréchet Inception Distance (FID) \cite{heusel2017gans}, LSE-C (Lip Sync Error - Confidence) \cite{prajwal2020lip} and LSE-D (Lip Sync Error - Distance) \cite{prajwal2020lip} are used as quantitative metrics. PSNR measures the pixel error between the original and reconstructed image \cite{kotevski2010experimental}. Higher PSNR means better quality. SSIM measures the similarity between the original and reconstructed image by comparing their structure, brightness and contrast \cite{wang2004image}. Higher SSIM means more similarity. LPIPS measures the perceptual similarity between two images by comparing their features from a deep network  \cite{zhang2018perceptual}. It aims to quantitatively reflect how humans perceive images \cite{zhang2018perceptual}. FID measures the realism of generated images by comparing their distributions with ground-truth images using their features from an Inception network \cite{heusel2017gans}. Lower scores for LPIPS and FID mean better quality. LSE-C and LSE-D are based on SyncNet \cite{Chung16a}, which is a lip-sync scorer. 
%We finetune DFRF \cite{shen2022dfrf} for 5000 steps on each speaker in the test set for a fair comparison.
\begin{figure*}[!ht]
  \centering
  \setlength{\belowcaptionskip}{-10pt}
  \includegraphics[width=1\textwidth]{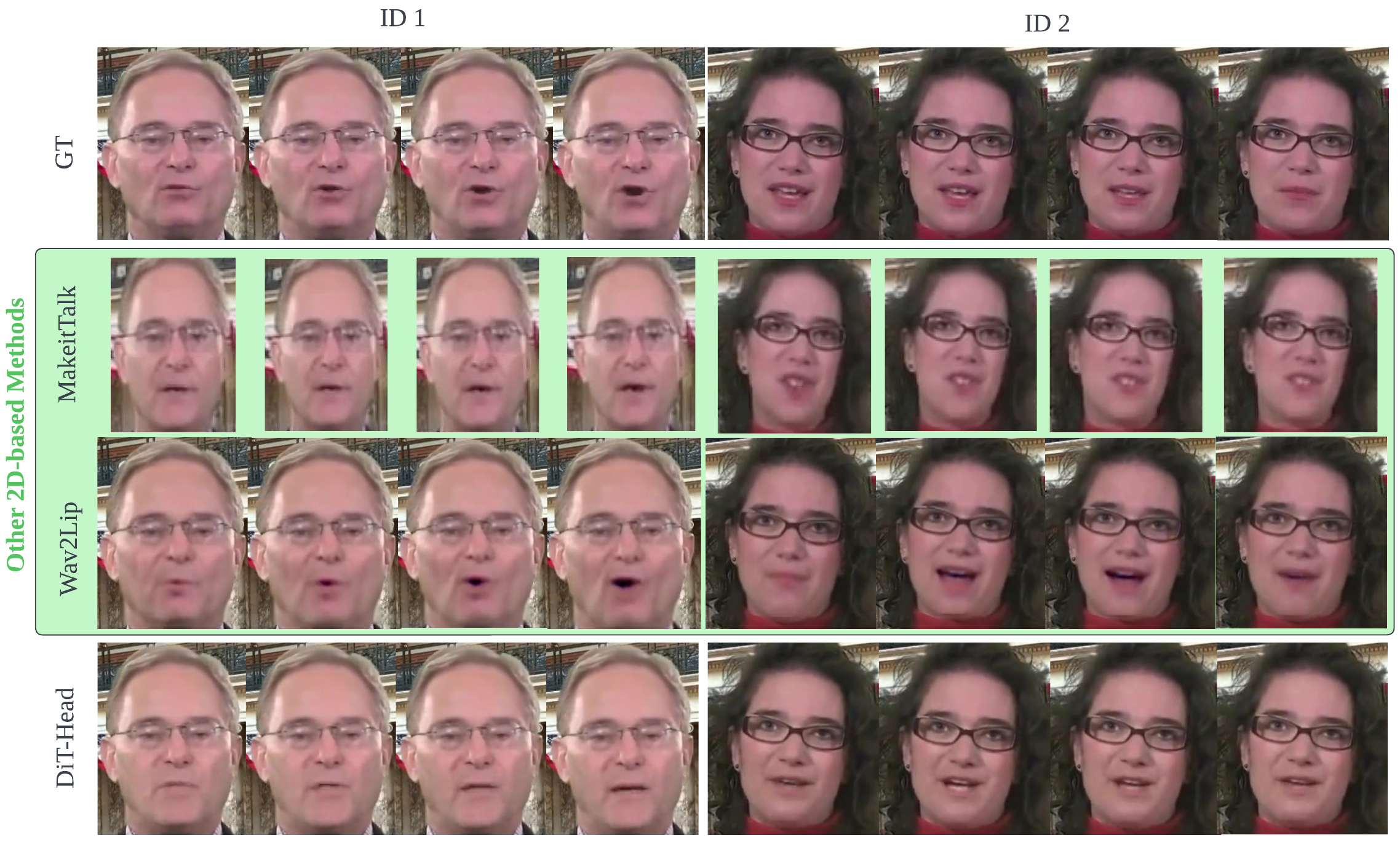}
  \caption{Qualitative comparison between DiT-Head and other 2D-based talking head synthesis methods (MakeItTalk \cite{zhou2020makelttalk}, Wav2Lip \cite{prajwal2020lip}). We encourage readers to view the supplementary video for a more accurate reflection of the qualitative differences between approaches.}
\end{figure*}

We compared padded face-crop videos from DiT-Head to padded face-crop videos from other talking head synthesis methods such as MakeitTalk \cite{zhou2020makelttalk} and Wav2Lip \cite{prajwal2020lip}. This reduces background bias but adds a padded area for blending assessment. MakeitTalk \cite{zhou2020makelttalk} and Wav2Lip \cite{prajwal2020lip} were chosen as they are among the current state-of-the-art for 2D-based person-agnostic talking head synthesis with public implementations available. These implementations were used to generate the output videos. Table 1 presents our quantitative findngs, which includes a quantitative comparison between MakeItTalk \cite{zhou2020makelttalk}, Wav2Lip \cite{prajwal2020lip} and DiT-Head w/o Stage 3. We use DiT-Head w/o Stage 3 for a fair comparison between frames and because the lip-sync scorer \cite{Chung16a} used for LSE-C\cite{prajwal2020lip} and LSE-D \cite{prajwal2020lip} metrics operates only on 25 FPS videos. Our qualitative findings are presented in Figure 2, which provides a visual comparison between MakeItTalk \cite{zhou2020makelttalk}, Wav2Lip \cite{prajwal2020lip} and DiT-Head.  

\begin{table*}[!ht]
<blank line>\vspace{5.247mm}
    \begin{center}
        \begin{tabular}{||c | c | c | c | c | c | c ||}
         \hline
         \textbf{Method} & \textbf{PSNR} $\uparrow$  & \textbf{SSIM} $\uparrow$ & \textbf{LPIPS} $\downarrow$ & \textbf{FID} $\downarrow$ & \textbf{LSE-C} $\uparrow$ & \textbf{LSE-D} $\downarrow$ \\ [0.5ex] 
         \hline
         MakeItTalk \cite{zhou2020makelttalk} & 22.47 & 0.73 & 0.207 & 55.20 & 3.40 & 11.4 \\ 
         \hline
         Wav2Lip \cite{prajwal2020lip} & 27.09 & 0.847 & 0.104 & 15.37 & \textbf{10.78} & \textbf{5.32} \\
         \hline
         % DFRF \cite{shen2022dfrf} & - & - & - & - & - & \cmark\\  
         % \hline
         \textbf{DiT-Head w/o Stage 3} & \textbf{28.37} & \textbf{0.872} & \textbf{0.0856}  & \textbf{10.31} & 3.60 & 10.49\\  %[1ex] 
         \hline 
         % \textbf{Ours} & - & - & - & - & - \\ [1ex] 
         % \hline
        \end{tabular}
    \end{center} 
\setlength{\belowcaptionskip}{-10pt}
\caption{Quantitative comparison on the test set between other 2D-based talking head synthesis methods (MakeItTalk \cite{zhou2020makelttalk}, Wav2Lip \cite{prajwal2020lip}) and DiT-Head w/o Stage 3. DiT-Head w/o Stage 3 was used for a fair comparison between frames. The best performance is highlighted in \textbf{bold}.}
\end{table*}

\subsection{Quantitative Comparisons with Other Methods}
Among the three methods, Table 1 shows that Wav2Lip \cite{prajwal2020lip} performs better than DiT-Head in lip-sync, as it has a lower LSE-D \cite{prajwal2020lip}, which measures the lip shape error on the whole region. DiT-Head w/o Stage 3 has a higher PSNR \cite{kotevski2010experimental} and SSIM \cite{wang2004image} in addition to lower LPIPS \cite{zhang2018perceptual} and FID \cite{heusel2017gans} than Wav2Lip. However, DiT-Head also has a lower LSE-C \cite{prajwal2020lip}, which measures the lip shape error on the center region. MakeItTalk \cite{zhou2020makelttalk} has the lowest PSNR \cite{kotevski2010experimental}, SSIM \cite{wang2004image} and LSE-C \cite{prajwal2020lip} as well as the highest LSE-D \cite{prajwal2020lip}, LPIPS \cite{zhang2018perceptual} and FID \cite{heusel2017gans} among the three methods. 
%\newpage
\subsection{Qualitative Comparisons with Other Methods}

Figure 2 shows that MakeItTalk \cite{zhou2020makelttalk} fails to preserve the pose of the input video and results in 
incorrect head movements and facial alignment, which are especially noticeable in the second identity. MakeItTalk \cite{zhou2020makelttalk} also produces outputs with 
little expression, as it cannot capture the subtle changes in the facial muscles that convey emotions. On the other hand, Wav2Lip \cite{prajwal2020lip} achieves the most accurate lip shape and expression among the three methods. However, Wav2Lip \cite{prajwal2020lip} suffers from low quality and blurriness around the mouth region. Moreover, Wav2Lip \cite{prajwal2020lip} generates a bounding box around the lower-half of the face, which can be seen as a sharp edge in some cases. This is more noticeable in Wav2Lip \cite{prajwal2020lip} on the second identity compared to DiT-Head. DiT-Head can generate high-resolution outputs with smooth transitions between the generated and original regions. However, DiT-Head has less accurate lip shape and expression than Wav2Lip \cite{prajwal2020lip}.  Furthermore, it can be seen that both Wav2Lip \cite{prajwal2020lip} and DiT-Head produce less fine-detailed texture of the face when compared to the ground-truth however the lip colour is more accurate in Wav2Lip \cite{prajwal2020lip}. We recognise that lip-sync quality is highly subjective. Hence, we encourage readers to view videos in our supplementary material (http://bit.ly/48MgiEr) for a more accurate reflection of the qualitative differences and the findings of our user study on visual quality, lip-sync quality and overall quality of DiT-Head compared to MakeItTalk \cite{zhou2020makelttalk} and Wav2Lip \cite{prajwal2020lip}.

\section{Discussion}

We aimed to compare different methods for talking head synthesis and to propose a novel method that can generate high-resolution and realistic outputs. The quantitative and qualitative results show that our model is competitive against other methods in terms of quality, similarity and lip shape accuracy.

Table 1 suggests that DiT-Head achieves the highest PSNR \cite{kotevski2010experimental} and SSIM \cite{wang2004image} compared to other methods, indicating 
the highest fidelity and similarity to the input video, as well as the lowest LPIPS \cite{zhang2018perceptual} and FID \cite{heusel2017gans}, indicating the lowest perceptual and Fréchet distance from the input video. This suggests that DiT-Head preserves the identity and expression of the ground-truth better than the other methods. However, Wav2Lip \cite{prajwal2020lip} performs the best in both LSE-C \cite{prajwal2020lip} and LSE-D \cite{prajwal2020lip}, indicating the most accurate lip shape on the center region and a similar lip shape on the whole region. This can be explained by the powerful lip-sync discriminator used in Wav2Lip \cite{prajwal2020lip} that ensures realistic lip movements and a perceptual loss that preserves the expression of the input. In contrast, DiT-Head outperforms MakeItTalk \cite{zhou2020makelttalk} in both LSE-C \cite{prajwal2020lip} and LSE-D \cite{prajwal2020lip} and uses only a reconstruction loss with no lip-sync discriminator which trains in a more stable manner and is not prone to mode collapse.

Although we recognise lip-sync quality is highly subjective, the qualitative results in Figure 2 also highlight the benefits of our model over the existing methods. Our model can generate high-resolution outputs with smooth transitions between the generated and original regions, as it uses frame-interpolation \cite{huang2020rife} and a temporal audio window to enforce temporal context. This enhances the quality and sharpness of the face. Our model can also capture the pose and expression of the input video better than MakeItTalk \cite{zhou2020makelttalk}, which uses a single identity image and audio to drive the talking head. Therefore, it cannot capture the identity and expression of the input as well as the other methods, nor can it generate realistic lip movements. Moreover, our model can avoid the blurriness and artifacts that affect Wav2Lip \cite{prajwal2020lip}, which is trained on low-resolution images and generates a bounding box around the lower-half of the face. Both Wav2Lip \cite{prajwal2020lip} and DiT-Head produce less fine-detailed texture of the face when compared to the ground-truth which shows that both methods may lose some high-frequency information in the face.

\subsection{Limitations and Future Work}

Despite the promising results of our deep learning model for talking head synthesis, we acknowledge there are some limitations that need to be addressed in future work. A drawback of our work is that our model employs a ViT model \cite{dosovitskiy2020image,vaswani2017attention} that requires a lot of 
computational resources and training time to achieve its high performance for multi-modal learning. Furthermore, we only train using a pre-trained audio model 
\cite{baevski2020wav2vec} for English speakers therefore DiT-Head cannot be effectively expanded to multiple languages.

Our model is scalable, but we aim to optimise our procedures to reduce the computational cost and speed up the training process by using flash-attention \cite{dao2022flashattention}. We also plan to explore temporal finetuning of the autoencoders to address the temporal jitter issue \cite{blattmann2023align}. Moreover, we acknowledge that the inference process for diffusion models is slower 
than GAN-based approaches, which is an open research problem for diffusion models. However, we still achieve a speedup compared to most person-specific 3D-based methods. For example, DFRF \cite{shen2022dfrf} takes about 130 hours to finetune on a specific speaker and 4 hours to render a 55-second video at 720x1280 pixel 
size at 25 FPS (using default settings, excluding pre-processing or post-processing) on a single RTX 3090, while DiT-Head can produce a talking-head video of the same speaker with any driving audio in 8 hours on the same hardware. We believe that by addressing these limitations, our model can become a viable approach for person-agnostic talking head synthesis that can generate realistic and expressive videos.

\subsection{Ethical Considerations}

Talking head synthesis models can create realistic videos of people speaking with any content, but they can also create “deepfakes” that manipulate or deceive others \cite{korshunov2022threat}. Deepfakes can harm individuals and groups in various ways and erode the trust and credibility of information sources. Thus, methods to detect, prevent, and regulate the misuse of talking head synthesis models are needed. To address this threat, users of DiT-Head for talking head generation will need to authenticate their credentials and watermark generated videos. We did not watermark our training dataset due to time constraints, but we plan to do so in future work to avoid deepfakes generated by users of our model.

\section{Conclusion}

We have introduced a novel and advanced solution for talking head synthesis, DiT-Head, that does not depend on the person’s identity and harnesses the powerful self- and cross-attention mechanisms of a DiT as opposed to UNets which suffer from a limited convolutional receptive field and GANs that suffer from unstable training and mode collapse. Our method outperforms existing ones in creating high-quality videos. The DiT enables our method to model the audio-visual-temporal dynamics of the input videos and produce realistic facial movements. In particular, the cross-attention mechanism allows our method to fuse audio and visual information for a more natural and coherent output. Our method opens up new possibilities for various applications in entertainment, education, and telecommunications. Of course, further work is needed to fully validate our technique: We are planning to extend the testing and validation of the model to faces from different ethnic groups (not just white ones) and to different languages (not only English) to avoid biases; and to expand the talking head synthesis and lip synchronization to more realistic cases such as blurred/low resolution images, occluded faces and real world scenarios involving not only straight, ID-type faces.

\bibliographystyle{apalike}
{\small
\bibliography{references}}

%\section*{\uppercase{Appendix}}

%If any, the appendix should appear directly after the
%references without numbering, and not on a new page. To do so please use %\textit{$\backslash$section*\{APPENDIX\}}

\end{document}